\documentclass[runningheads]{llncs}

\usepackage{eccv}

\usepackage{eccvabbrv}

\usepackage{graphicx}
\usepackage{booktabs}
\usepackage{xcolor,colortbl}
\usepackage{comment}
\usepackage[accsupp]{axessibility}  
\definecolor{lightgray}{rgb}{0.88, 0.88, 0.88}

\usepackage{hyperref}

\usepackage{orcidlink}
\usepackage{graphicx}
\usepackage{amsmath}
\usepackage{amssymb}
\usepackage{booktabs}
\usepackage{multirow}
\usepackage{tabularx}
\usepackage{bm}
\usepackage{wrapfig}
\newcolumntype{g}{>{\columncolor{Gray}}c}
\usepackage{physics}
\newcolumntype{C}{>{\centering\arraybackslash}X} 

\begin{document}

\title{Multimodal Cross-Domain Few-Shot Learning \\ for Egocentric Action Recognition} 

\titlerunning{MM-CDFSL}

\author{Masashi Hatano\inst{1}\orcidlink{0009-0002-7090-6564} \and
Ryo Hachiuma\inst{2}\orcidlink{0000-0001-8274-3710} \and
Ryo Fujii\inst{1}\orcidlink{0000-0002-9115-8414} \and
Hideo Saito\inst{1}\orcidlink{0000-0002-2421-9862}}

\authorrunning{M.~Hatano et al.}

\institute{
Keio University \and
NVIDIA
}

\maketitle

\begin{abstract}
We address a novel cross-domain few-shot learning task (CD-FSL) with multimodal input and unlabeled target data for egocentric action recognition.
This paper simultaneously tackles two critical challenges associated with egocentric action recognition in CD-FSL settings: (1) the extreme domain gap in egocentric videos (\eg, daily life vs. industrial domain) and (2) the computational cost for real-world applications.
We propose MM-CDFSL, a domain-adaptive and computationally efficient approach designed to enhance adaptability to the target domain and improve inference cost.
To address the first challenge, we propose the incorporation of multimodal distillation into the student RGB model using teacher models.
Each teacher model is trained independently on source and target data for its respective modality.
Leveraging only unlabeled target data during multimodal distillation enhances the student model's adaptability to the target domain.
We further introduce ensemble masked inference, a technique that reduces the number of input tokens through masking. 
In this approach, ensemble prediction mitigates the performance degradation caused by masking, effectively addressing the second issue.
Our approach outperformed the state-of-the-art CD-FSL approaches with a substantial margin on multiple egocentric datasets, improving by an average of 6.12/6.10 points for 1-shot/5-shot settings while achieving $2.2$ times faster inference speed.
Project page: \url{https://masashi-hatano.github.io/MM-CDFSL/}

 \and Multimodal Distillation
    
\end{abstract}

\section{Introduction}

The field of egocentric vision research, primarily facilitated by wearable devices such as smart glasses~\cite{aria}, has seen significant development over the last decades, owing to its wide range of application domains, including daily life, industry, AR/VR, and medicine.
In response to the growing demand across multiple domains, several large-scale egocentric datasets, such as Ego4D~\cite{ego4d} and Ego-Exo4D~\cite{ego-exo4d}, have been proposed to provide a variety of research tasks with various modalities~\cite{gong2023mmg, Plizzari_2022_CVPR, Wang_2023_CVPR, Huang_2023_CVPR, Ryan_2023_CVPR, spotem}. 
Recognizing the action of the camera wearer (\textit{egocentric action recognition}) is one of the fundamental tasks in the egocentric video understanding domain.

Despite these advancements, a scarcity of egocentric datasets persists in various domains, such as industry~\cite{meccano, schoonbeek2024industreal, ragusa2023enigma51, sener2022assembly101} and the medical field~\cite{wang2023pov, Fujii2024EgoSurgeryPhase}.
Cross-domain approaches~\cite{tzeng2017adversarial, bousmalis2017unsupervised, long2018conditional} effectively transfer the knowledge acquired from the training on large-scale datasets (\ie, source domain) to the target domain. 
Since the actions taken in the target domain may differ from those in the source domain, the time-consuming manual annotation of the action labels is required.
Therefore, cross-domain few-shot learning (CD-FSL)~\cite{Liang_2021_ICCV, Li_2022_CVPR, hu2022adversarial, Zhou_2023_CVPR, Zhao_2021_WACV} emerges as a crucial strategy, merging cross-domain \textit{adaptability} with the efficiency of few-shot learning~\cite{NeurIPS2017Shell, NeurIPS2016ShellVinyals, ICML2017Chelsea, Lee_2019_CVPR, sung2018learning, NeurIPS2018ZHANG} which only uses a few labeled samples on the target domain.
Recently, using unlabeled data on the target domain has proven effective in enhancing the adaptability against the domain without increasing the annotation cost~\cite{startup, dynamic-distill, cdfsl-v}.
These approaches further improve the task performance in the CD-FSL setting.

\begin{figure}[t]
\begin{center}
   \includegraphics[width=1\linewidth]{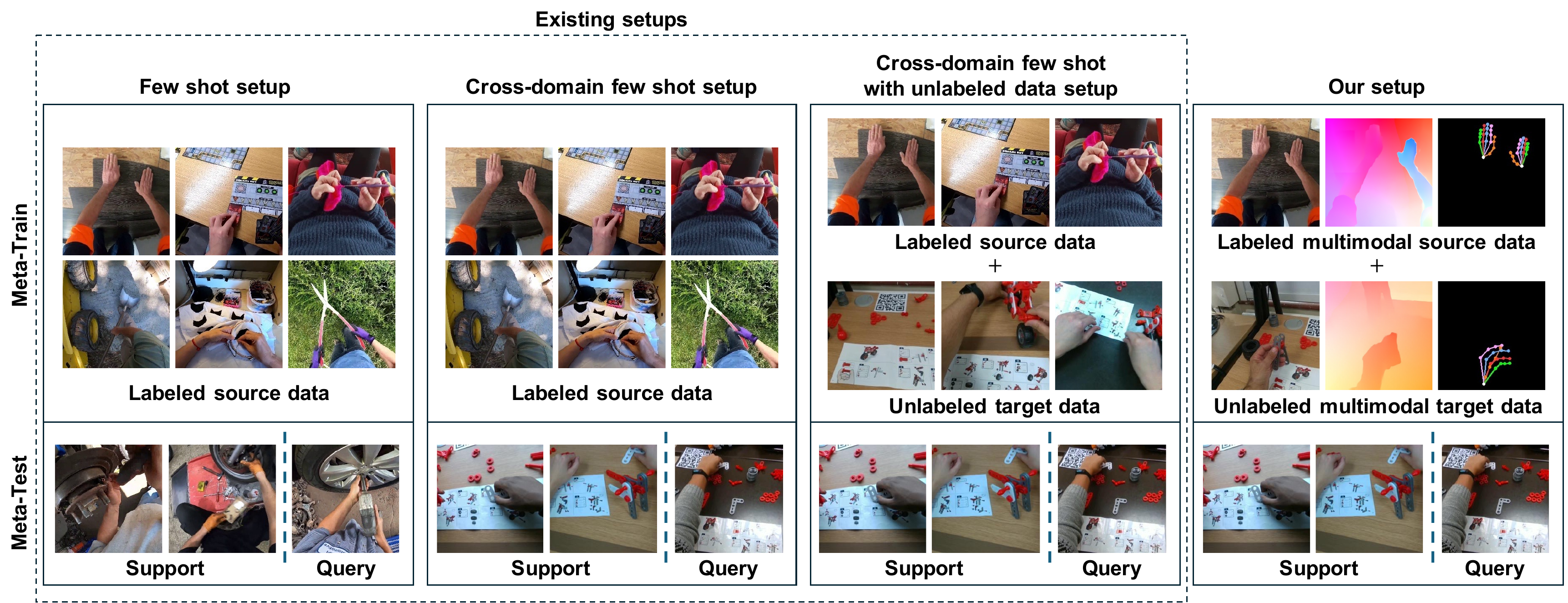}
\end{center}
   \caption{\textbf{Problem setup.} In our problem setup, a model is trained using source data and unlabeled target data for multiple modalities during the meta-training stage. In the meta-testing stage, a few examples of novel classes from the support set are provided to learn a classifier. Then, the network predicts the categories of different samples from the query set, which are the same classes as the support set. Unlike existing setups, we leverage multimodal data (\eg, optical flows or hand poses) during the meta-training stage. During the meta-testing stage, only RGB videos are used as inputs.
   }
\label{fig: problem_setting}
\end{figure}

The CD-FSL with unlabeled target data~\cite{startup, dynamic-distill, cdfsl-v} adopts two meta-training stages (pretraining and domain adaptation) and two meta-testing stages (few-shot training and inference).
For example, Dynamic Distillation~\cite{dynamic-distill} first trains the visual encoder on the source dataset (pretraining stage), subsequently training with labeled source and unlabeled target data via pseudo distillation~\cite{xie2020self, zhang2021prototypical} (domain adaptation stage). 
Lastly, a classifier is trained on the few-shot support data of novel classes (few-shot training stage) and evaluated on the query samples (inference).
Building upon this, CDFSL-V~\cite{cdfsl-v} proposed to use the recent self-supervised technique, VideoMAE~\cite{videomae}, for the pretraining stage, enabling the pretraining with the unlabeled target data to enhance the adaptability to the target domain on the action recognition task.

Although the CD-FSL task with unlabeled target data has been widely investigated, two significant issues remain when applying the previous works~\cite{startup, dynamic-distill, cdfsl-v} to the egocentric action recognition task.

\noindent \textbf{Adaptability to the Target Domain.}
Solely relying on visual information is susceptible when performing on the target domain despite using domain adaptation techniques. 
While the previous works have focused only on RGB modality for image or video classification tasks, it is known that using multimodal information (\eg, optical flows or poses) helps to mitigate the domain gap between the source and target domains for the action recognition task~\cite{Moon2021CVPR,Munro2020CVPR}. 
Especially on the egocentric action recognition task, since the background's visual information vastly changes, solely relying on the RGB information to adapt the domain is infeasible.

\noindent \textbf{Inference Cost.}
Despite the advancements in the CD-FSL task expanding for videos~\cite{cdfsl-v}, processing densely-sampled input frames with temporal-aware operations is necessary, making the process computationally expensive~\cite{Kondratyuk2021CVPR,Junke2022ECCV,wang2023maximizing}.
This computational intensity hampers the practical applications on edge devices with limited resources.
Therefore, reducing the inference cost is essential for egocentric action recognition.

In this work, for the first time, we study cross-domain few-shot learning for egocentric action recognition using multimodal input and unlabeled target data (\cref{fig: problem_setting}).
We propose MM-CDFSL, a novel CD-FSL approach utilizing multimodal input and unlabeled target data to enhance the adaptability to the target domain.
In our meta-training framework, we first pretrain the domain-adapted and class-discriminative features at the pretraining stage.
Subsequently, we conduct multimodal distillation to bridge the domain gap further.
This work capitalizes on incorporating multiple modalities to alleviate the extreme domain gaps on the CD-FSL task.
Additionally, we consider the computational cost during inference by reducing the number of input tokens without compromising the action recognition accuracy.
We aim to simultaneously achieve strong adaptability to the target domain where the shift between the source and target domain is significant while improving the runtime during the inference.

To address the first challenge, we propose to (1) incorporate supervised training on source-labeled data in the pretraining stage and (2) additionally introduce the multimodal distillation stage to transfer the knowledge obtained during the pretraining to a network with RGB input.
In the pretraining stage, domain-adapted and class-discriminative features are obtained using labeled source and unlabeled target data.
Utilizing VideoMAE~\cite{videomae} for reconstructing source and target inputs, our model aims to capture discriminative and shared representations between source and target domains.
This pretraining process is applied independently for all input modalities.
Subsequently, we perform multimodal feature distillation on the unlabeled target dataset to transfer knowledge from the teacher models trained on source and target domains for multiple modalities to the student RGB encoder.
Since the teacher networks are trained on both source and target data, leveraging only unlabeled target data during multimodal distillation helps improve the adaptability of the student network to the target domain.

To address the challenge of computational efficiency, we propose an \textit{Ensemble Masked Inference}, which reduces the number of input tokens by randomly masking tokens from the input data and then ensembling multiple classification results estimated from these inputs.
However, naively masking the input data during inference can lead to a severe performance drop since the feature extractor and the classifier are trained on the unmasked inputs. Therefore, drawing inspiration from the \textit{Tube Masking} operation in VideoMAE~\cite{videomae}, originally proposed to randomly mask input tokens in videos to obtain meaningful representations during pretraining, we propose to apply this operation across all stages, including pretraining, multimodal distillation, few-shot training, and inference. By applying masking throughout these stages, we prevent distribution drift while simultaneously improving inference speed.

In summary, our contributions are as follows:
\begin{itemize}
    \item We propose a novel, challenging, but practical problem: cross-domain few-shot learning with multimodal input and unlabeled target data in egocentric scenarios.
    \item We propose MM-CDFSL, a novel approach for the CD-FSL for egocentric action recognition task that utilizes a domain-adapted and class-discriminative pretraining and multimodal feature distillation.
    Furthermore, we propose ensemble masked inference to reduce the computational cost.
    \item We simultaneously achieve state-of-the-art performance in terms of the accuracy and inference speed on multiple egocentric action recognition benchmarks~\cite{epic,meccano,wear} with the CD-FSL settings. Specifically, our method outperforms the prior state-of-the-art in accuracy by an average of 6.12/6.10 points for 1-shot/5-shot settings, while our approach is 2.2x faster than the previous approaches.
\end{itemize}

\section{Related Work}
First, we discuss the most relevant work on egocentric action recognition and then review previous efforts on cross-domain few-shot learning (CD-FSL).
Our work intersects with prior research from several perspectives: few-shot learning, cross-domain adaptability, the utilization of unlabeled target data, and the incorporation of multimodal inputs. 
To provide a clear and concise comparison, we have summarized the differences in \cref{tab:survey-table}.

\noindent \textbf{Egocentric Action Recognition}.
Egocentric action recognition has gained popularity~\cite{hamed2012cvpr, tekin2019cvpr, poleg2016wacv, Wang_2021_ICCV} with the advent of affordable, lightweight wearable cameras, such as GoPro. 
This increase in egocentric data has sparked significant interest in understanding and recognizing actions within these videos~\cite{ear-survey}.
One of the key challenges for egocentric action recognition is an environmental bias where egocentric videos are captured in different locations.
For example, some researchers~\cite{Song_2021_CVPR, Plizzari2023iccv} have aimed to acquire a domain generalized or domain adapted representation.
Also, the lack of large-scale datasets with annotations is a fundamental problem in egocentric vision, especially in the industrial and medical fields.
This necessitates the few-shot learning technique, which enables the inference of novel classes with limited labeled samples, thereby fueling research in this domain~\cite{gong2023mmg, wang2022hybrid}.
In contrast to the previous work, we simultaneously tackle these two key challenges.

On the other hand, incorporating additional multimodal information is essential for identifying activities for egocentric activity recognition. 
Several studies have demonstrated that leveraging supplementary modalities at inference time significantly enhances performance~\cite{kazakos2021MTCN,kazakos2019TBN,NEURIPS2021_76ba9f56,Radevski_2023_ICCV,WangECCV2018,Materzynska2020CVPR,gabeur2020mmt}. 
However, acquiring and processing the additional modalities typically demands significant computational resources. 
This limitation renders these methods cumbersome or impractical, particularly on constrained computing budgets such as embedded devices. 
To benefit from multimodal information without increasing computational cost at inference time, the distillation technique has attracted numerous researchers~\cite{tan2023egodistill, Radevski_2023_ICCV}.
Following this success, our approach incorporates multimodal distillation.
However, unlike the existing work~\cite{Radevski_2023_ICCV, tan2023egodistill}, we focus on a challenging but practical scenario: cross-domain and few-shot settings for egocentric action recognition.

\begin{table}[tb]
\caption{Comparative overview of methodological features.}
\centering
\begin{tabular}{c|c|c|c|c}
\toprule
Methods & Few-shot & Cross-domain & Unlabeled target data & Multimodal \\
\hline
 \cite{NeurIPS2016ShellVinyals,NeurIPS2017Shell,ICML2017Chelsea,NeurIPS2018ZHANG,sung2018learning,Lee_2019_CVPR} & \checkmark &   & &     \\
\cite{bscd-fsl, tseng2020iclr, Wang_2022_CVPR, Liu_2021_ICCV} & \checkmark & \checkmark  & &   \\
\cite{startup, dynamic-distill, cdfsl-v} & \checkmark & \checkmark & \checkmark &  \\
\cite{Radevski_2023_ICCV, tan2023egodistill} & & & & \checkmark   \\
\cite{gong2023mmg} & \checkmark & & & \checkmark \\
\rowcolor{lightgray}
MM-CDFSL (Ours) & \checkmark & \checkmark & \checkmark & \checkmark\\
\bottomrule
\end{tabular}
\label{tab:survey-table}
\end{table}

\noindent \textbf{Cross-Domain Few-Shot Learning}.
Guo \etal~\cite{bscd-fsl} introduced the Cross-Domain Few-Shot Learning (CD-FSL) challenge, which assesses the capability of deep neural networks to adapt image classification to novel domains that present classes not encountered during their training. 
Research has explored numerous strategies, such as fine-tuning~\cite{chen2019closerfewshot}, feature-wise transformation~\cite{tseng2020iclr}, prototype learning~\cite{Wang_2022_CVPR}, and the training with multiple domains~\cite{Liu_2021_ICCV}, all aimed at enhancing the networks' generalization capabilities. 
Among these strategies, using unlabeled target data stands out for its ability to increase adaptability to target domains, offering a practical approach akin to unsupervised domain adaptation~\cite{hoffman2018cycada, ganin2015unsupervised}, as the annotation for large-scale target data is not required.

To incorporate unlabeled target data during training, STARTUP~\cite{startup}
and Dynamic Distillation~\cite{dynamic-distill} employ a pseudo-labeling technique based on the self-training strategy.
Both methods first pretrain the visual feature extractor on the labeled source dataset, subsequently training with student-teacher networks initialized from the pretrained weights.
CDFSL-V~\cite{cdfsl-v} addresses challenging cross-domain few-shot learning for action recognition.
The method incorporates the recent self-supervised technique for video understanding, VideoMAE~\cite{videomae}, to utilize unlabeled target data during the pretraining stage, leading to better adaptability to the target domain.

However, neither of these approaches investigates the leverage of multiple modalities, which is crucial in egocentric action recognition, and considers inference time.
In contrast, we explore utilizing multiple modalities for adaptability to the target domain and consider the computation cost.
\section{Method}
\begin{figure}[t]
\begin{center}
   \includegraphics[width=\linewidth]{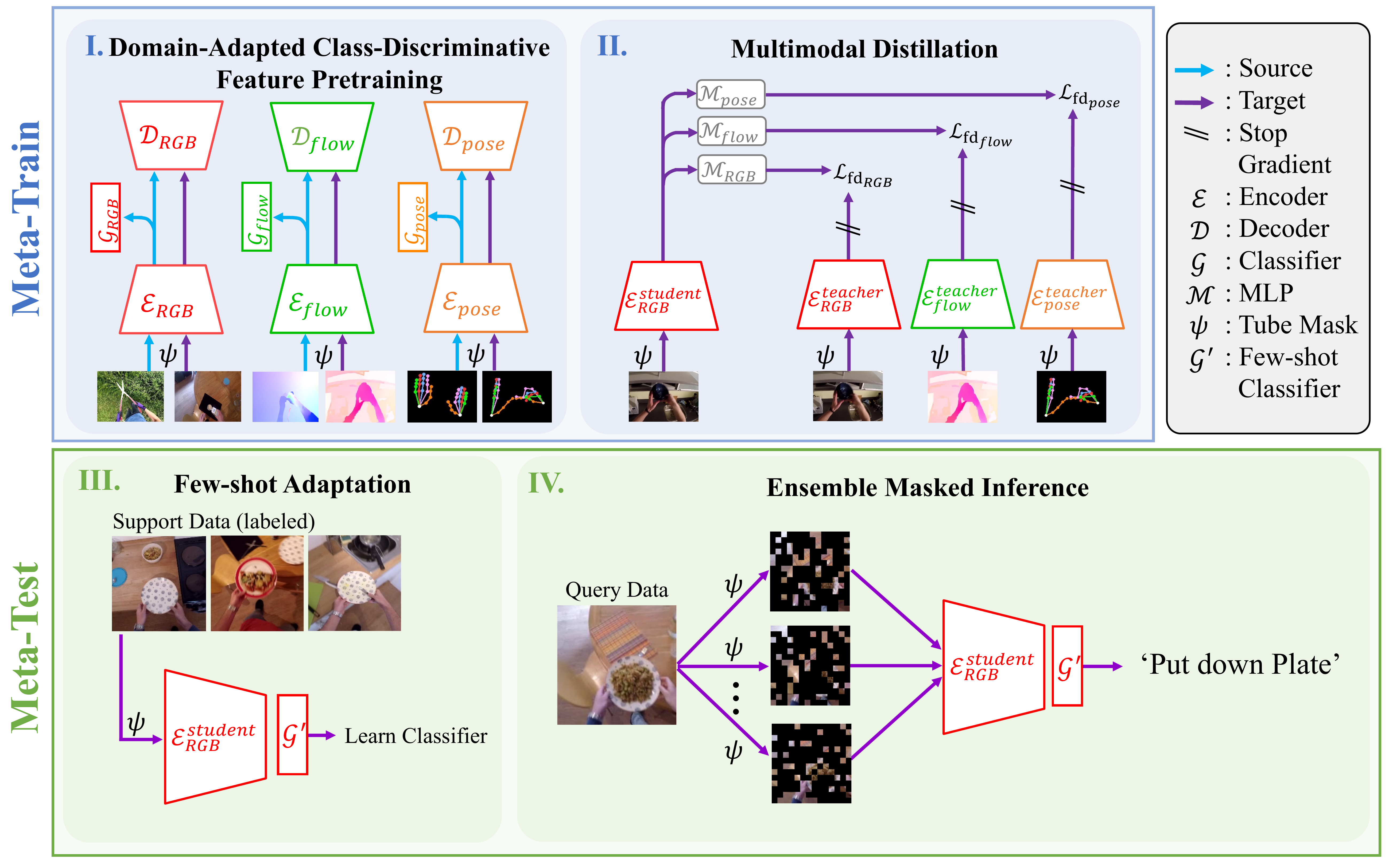}
\end{center}
   \caption{\textbf{The framework of our proposed method.} Our approach has two meta-training and two meta-testing stages: 1. learning domain-adapted and class-discriminative features for all modalities, 2. distilling the multimodal features into student RGB encoders, 3. few-shot learning for adapting novel classes, and 4. ensemble masked inference using $P$ Tube Masking during inference.}
\label{fig: overview}
\end{figure}

The proposed method consists of two meta-training and two meta-testing stages: the first involves domain-adapted and class-discriminative feature pretraining, the second benefits the domain adaptability from multimodal distillation, the third trains a classifier to adapt novel classes using labeled few-shot samples, and the fourth infer the action class of query data while reducing the inference time.
We first introduce a new problem setup of the cross-domain few-shot learning task with multimodal input and unlabeled target data (\cref{sec:problem-definition}).
Then, we introduce the proposed method, including (1) pretraining (\cref{sec:pretraining}), (2) multimodal distillation (\cref{sec:mmdistill}), (3) few-shot training and (4) ensemble masked inference (\cref{sec:inference}).
\cref{fig: overview} provides an overview of our approach.
For readability, the following section is explained with RGB, optical flow, and heatmap of the hand pose (referred to as \textit{hand pose}) as multimodal information, known as effective modal information for egocentric action recognition. However, our method can use any modality information (\eg, IMU, audio).

\subsection{Problem Definition}
\label{sec:problem-definition}
The goal of the cross-domain few-shot learning (CD-FSL) task with multimodal input and unlabeled target data is to classify novel classes within the target dataset $D_{T}$, leveraging both a labeled source dataset $D_{S}$ and an unlabeled target dataset $D_{T_{u}}$. 
Both $D_{S}$ and $D_{T_{u}}$ comprise data across $m$ modalities. 
Ensure there is no class overlap between the source and target datasets to maintain the integrity of the few-shot learning scheme.
To infer novel classes in the target dataset, the data $D_{T_{u}}$ is split into a support set $S$, which contains $K$ data from $N$ classes for the $N$-way $K$-shot task, and a query set $Q$ consisting of data from only the $N$ classes of the support set, same as the standard few-shot learning setting.

\subsection{Domain-adapted and Class Discriminative Feature Pretraining}
\label{sec:pretraining}
In the pretraining stage, the VideoMAE~\cite{videomae} for each modal is independently trained to learn the representation shared among the source/target domain and discriminative features on the source domain via two objective functions: (1) joint reconstruction of the source and target domain data and (2) the classification of the action categories on the source dataset. In contrast to the pretraining stage in CDFSL-V~\cite{cdfsl-v}, which only reconstructs the data on the source and target dataset to mitigate the domain shift, we train the model via a joint optimization of the above two objectives to effectively learn the shared and discriminative feature representation.

Specifically, given an input $\bm{x}_{m} \in \mathbb{R}^{T \times H_{m} \times W_{m} \times C_{m}}$ consisting $T$ frames of image size $H_{m} \times W_{m}$ with a channel size $C_{m}$ for modality $m \in \{$RGB, optical flow, hand pose heatmap$\}$, respectively, we train a VideoMAE~\cite{videomae}, which consists of encoder (feature extractor) $\mathcal{E}_{m}$ and decoder $\mathcal{D}_{m}$, per modality $m$. Following the training framework of VideoMAE, the tokenized input data are randomly masked with the masking ratio $\rho_{\text{pretrain}}$ by \textit{Tube Masking} $\psi$ and reconstructed as follows:
\begin{equation}
    \bm{\hat{x}}_{m} = \mathcal{D}_{m}(\mathcal{E}_{m}(\psi(\bm{x}_{m}))),
\end{equation}
where $\bm{\hat{x}}_{m}$ denotes reconstructed input.
Additionally, a classifier $\mathcal{G}_{m}$ is employed to input the averaged embedding extracted from $\mathcal{E}_{m}$ and estimates the logit $l_m \in \mathbb{R}^{n_{c}}$ ($n_c$ denotes the number of the classes in source dataset) as follows: $l_m = \mathcal{G}_{m}(\mathcal{E}_{m}(\psi(\bm{x}_{m})))$.

We train $\mathcal{E}_{m}$, $\mathcal{D}_{m}$, and $\mathcal{G}_{m}$ for each modal with the combination of three losses: the reconstruction loss for source data $\mathcal{L}^{\scalebox{0.5}{source}}_{\text{recon}}$, the reconstruction loss for target data $\mathcal{L}^{\scalebox{0.5}{target}}_{\text{recon}}$, and the cross entropy loss for action classification in the source dataset $\mathcal{L}^{\scalebox{0.5}{source}}_{\text{ce}}$ with a balancing hyperparameter $\lambda_{\text{ce}_{m}}$:
\begin{equation}
\label{eq:pretrain-loss}
    \mathcal{L}_{\text{pretrain}} = \mathcal{L}^{\scalebox{0.5}{source}}_{\text{recon}} + \mathcal{L}^{\scalebox{0.5}{target}}_{\text{recon}} + \lambda_{\text{ce}_{m}} \mathcal{L}^{\scalebox{0.5}{source}}_{\text{ce}}.
\end{equation}
Note that we only calculate $\mathcal{L}^{\scalebox{0.5}{source}}_{\text{ce}}$ against labeled source data, and we employ $L_2$ loss for the reconstruction loss.

\subsection{Multimodal Distillation}
\label{sec:mmdistill}
Next, we further improve the adaptability to the target domain by distilling the domain-adapted multimodal features to the RGB feature extractor $\mathcal{E}_{\scalebox{0.5}{RGB}}$.
Using multiple modalities helps mitigate the domain shift between source and target domain compared to using only RGB modality, as visual information is susceptible to lighting, background, and appearance variations.
Incorporating additional modalities like optical flow and hand pose introduces complementary information that is less sensitive to these visual changes.
Furthermore, distilling the multimodal features, which are domain-adapted and class-discriminative, into the RGB modality reduces the model complexity while alleviating the domain gap.
It aims to imbue the RGB model with domain-adapted characteristics of multimodal learning without the overhead of processing and integrating multiple modality data types during inference.

Given multimodal input for unlabeled target data $\bm{x}^{\scalebox{0.5}{target}}_{m}$, we train the student RGB encoder $\mathcal{E}^{\scalebox{0.5}{student}}_{\scalebox{0.5}{RGB}}$ using the teacher encoder $\mathcal{E}^{\scalebox{0.5}{teacher}}_{m}$ for all modalities, including RGB.
Distilling features from the RGB modality ensures that it regularizes the multimodal distillation from other modalities to prevent forgetting the RGB element and biasing the other modalities.
It is noted that all student and teacher encoders are initialized from the pretrained weights in the previous domain-adapted and class-discriminative feature pretraining section, and the weights of teacher encoders are kept frozen during the multimodal distillation stage.
Also, we mask the input tokens like the previous self-supervised training, with the same mask ratio $\rho_{\text{distill}}$ across the modalities.
The masking is adopted as the model is required to make predictions based on masked inputs during inference (See \cref{sec:inference}).
The unlabeled target RGB data is fed to the student RGB encoder; subsequently, the extracted features are projected into modalities by projection layers (\eg, multilayer-perceptrons) $\mathcal{M}_{{m}}$:
\begin{equation}
    \bm{\hat{f}}_{m} = \mathcal{M}_{{m}}(\mathcal{E}^{\scalebox{0.5}{student}}_{\scalebox{0.5}{RGB}}(\psi(\bm{x}^{\scalebox{0.5}{target}}_{\scalebox{0.5}{RGB}}))),
\end{equation}
where $\bm{\hat{f}}_{m}$ are projected features from RGB to the modality $m$.

We aim to minimize its $L_{2}$ distance from the real embedding of the modality $m$. 
Specifically, the loss is computed as a linear combination of the $L_{2}$ losses corresponding to each modality, and the feature distillation loss is defined as follows:
\begin{equation}
\mathcal{L}_{\text{fd}_{m}} = \norm{ \text{sg}[\bm{f}_{m}] - \bm{\hat{f}}_{m} }_{2}^{2},
\end{equation}
where $\text{sg}[.]$ stands for the stop gradient operator that is defined as an identity at forward computation time and has zero partial derivatives, and $\bm{f}_{m}$ denotes the extracted features of the unlabeled target data $\bm{x}^{\scalebox{0.5}{target}}_{m}$ from the teacher encoder of the modality $m$:
\begin{equation}
    \bm{f}_{m} = \mathcal{E}^{\scalebox{0.5}{teacher}}_{m}(\psi(\bm{x}^{\scalebox{0.5}{target}}_{m})).
\end{equation}
The final training loss in the multimodal distillation stage is the linear combination of these losses for each modality:
\begin{equation}
\label{eq:mmdistill-loss}
    \mathcal{L}_{\text{distill}} = \sum_{m} \mathcal{L}_{\text{fd}_{m}}.
\end{equation}

\subsection{Ensemble Masked Inference}
\label{sec:inference}
The computation cost is one of the fundamental problems for a real-time application or inference on limited-resourced devices.
Processing all tokens in input frames for the Transformer model for action recognition is computationally expensive~\cite{Kondratyuk2021CVPR,Junke2022ECCV,wang2023maximizing}; however, the computation cost can be modulated depending on how many tokens from input frames are used. 
On the one hand, existing methods using the ViT architecture process all tokens from input frames to achieve strong action recognition performance, but this comes at a high computational cost.
The attention mechanism demands computational complexity of $\order{I^2}$ where $I$ denotes the number of input tokens.
On the other hand, reducing the number of input tokens helps mitigate the computation cost but results in a drop in performance.
To this end, we propose the ensemble masked inference to reduce the computation cost by reducing the number of input tokens by masking with the mask ratio $\rho_{\text{infer}}$ and alleviate the performance drop by utilizing the ensemble prediction with the ensemble number $P$: $\order{P((1-\rho_{\text{infer}})I)^2}$.

\noindent \textbf{Few-Shot Training}.
Following the existing work~\cite{startup, dynamic-distill, cdfsl-v}, we learn a new classifier for adapting the novel classes in the target domain with a few numbers of labeled data.
The RGB student encoder is retained, and the classifier head $\mathcal{G'}$ on top of the encoder is trained with a sampled $N$-way $K$-shot data from the support set $S$.
The Tube Masking is applied to the input with the mask ratio $\rho_{\text{infer}}$ during the few-shot training process, enabling the model to make predictions based on masked input at the inference time.
Note that the Tube Mask is varied across the input data from the support set to prevent the model from overly relying on identical masking patterns, which may inadvertently mask all-important content. 

\noindent \textbf{Inference}.
Once the classifier $\mathcal{G'}$ is adapted to the novel classes in the target domain, $q$ samples from each $N$ class from the query set $Q$ are used to evaluate the few-shot action recognition.
We apply the Tube Masking $\psi$ with the same mask ratio $\rho_{\text{infer}}$ used during the few-shot training.
Various mask ratios $\rho_{\text{infer}}$ are applied depending on the specific requirements for a trade-off between accuracy and inference speed.
Adjusting the mask ratio allows us to tailor the model's performance according to the computational constraints or the precision demands of the application at hand. 
Furthermore, we adopt ensemble learning to mitigate the drop in performance caused by masking the input frames. 
We generate $P$ data from one sample by applying varied Tube Masking and then average the predicted probabilities:
\begin{equation}
    \hat{y} = \frac{1}{P} \sum_{p} \text{Softmax}(\mathcal{G'}(\mathcal{E}_{\scalebox{0.5}{RGB}}(\psi(\bm{x}_{\scalebox{0.5}{RGB}})))).
\end{equation}

\section{Experiments}
In this section, we elaborate datasets used for training and evaluation, comparison methods, quantitative comparison, and ablation study of our proposed design. (See Suppl. for implementation details and more results).

\subsection{Datasets}
We employ the most large-scale egocentric video dataset, Ego4D~\cite{ego4d} as a source dataset and multiple egocentric datasets~\cite{epic,meccano,wear} as a target dataset.

\noindent \textbf{Ego4D~\cite{ego4d}}.
The Ego4D dataset is one of the large-scale egocentric video datasets in a daily life domain. 
It contains 3,670 hours of egocentric videos of people performing diverse tasks, such as gardening or crafting, and is collected by 931 people from 74 locations across nine different countries worldwide.
We utilize annotations for short-term action anticipation, including both the clip frame and the corresponding time to contact for the egocentric action recognition task. 
For our experiment, we constructed annotations by selecting clip frames, including the frame at which contact occurs within the input sequence.
This results in a $204$-class action recognition dataset with $15.5$k training clips.

\noindent \textbf{EPIC-Kitchens~\cite{epic}}.
EPIC-Kitchens is a dataset captured in the cooking domain, where $32$ participants have recorded $432$ egocentric videos.
The dataset encompasses a total of $286$ action classes.
For our cross-domain few-shot study, we partitioned these videos into training (unlabeled) and validation subsets, comprising $58$ and $228$ action classes with $3.6$k and $17.2$k video clips, respectively.
To adhere to the few-shot learning task settings, we ensured no class overlap between the training (unlabeled) and validation sets.

\noindent \textbf{MECCANO~\cite{meccano}}.
The MECCANO dataset was collected in an industrial-like setting, where $20$ participants were asked to assemble a toy model of a motorbike.
We partitioned the MECCANO dataset into training and validation subsets.
This partition yielded $10$ and $40$ action classes, accompanied by $1.3$k and $7.4$k action segments for the training and validation sets.

\noindent \textbf{WEAR~\cite{wear}}.
The WEAR dataset, curated explicitly for human activity recognition within the outdoor sports domain, features egocentric video data capturing a wide array of athletic activities in natural settings.
It encompasses recordings from $18$ participants engaging in $18$ distinct workout activities across ten diverse outdoor locations.
To facilitate our study, we divided the dataset into training and validation subsets, allocating $3$ and $15$ activity classes to each, accompanied by $1.8$k and $8.0$k activity segments, respectively.

\begin{table}[tb]
\caption{\textbf{Cross-domain few-shot action recognition accuracy}. We assess the performance of 5-way 1-shot and 5-shot top-1 action recognition accuracy on three egocentric datasets, EPIC-Kitchens (EPIC), MECCANO (MEC), and WEAR. The Ego4D dataset is used as the source dataset. We report an average of 600 runs of few-shot evaluation with 95$\%$ confidence interval. We present the results of our method that adopts the mask ratio $\rho_{\text{infer}} = 0.75$ and ensemble number $P = 2$. The best values are shown in \textbf{bold}.}
\centering
\resizebox{\textwidth}{!}{
\begin{tabular}{lcccccc}
\toprule
\multirow{2.5}{*}{Method} & 
\multicolumn{3}{c}{1-shot} & 
\multicolumn{3}{c}{5-shot} 
\\
\cmidrule(l{2pt}r{3pt}){2-4} \cmidrule(l{2pt}r{3pt}){5-7} 
& EPIC & MEC & WEAR & EPIC & MEC & WEAR \\
\toprule
Random Initialization & 29.20\scalebox{0.7}{$\pm$.37} & 23.10\scalebox{0.7}{$\pm$.24} & 25.96\scalebox{0.7}{$\pm$.27} &  40.28\scalebox{0.7}{$\pm$.42} &  27.04\scalebox{0.7}{$\pm$.28} & 38.71\scalebox{0.7}{$\pm$.36} \\
VideoMAE~\cite{videomae} & 35.07\scalebox{0.7}{$\pm$.41} & 27.75\scalebox{0.7}{$\pm$.31} & 44.65\scalebox{0.7}{$\pm$.38} &  47.13\scalebox{0.7}{$\pm$.43} & 35.92\scalebox{0.7}{$\pm$.33} & 63.92\scalebox{0.7}{$\pm$.35} \\
STARTUP\scalebox{0.7}{++}~\cite{startup} & 35.18\scalebox{0.7}{$\pm$.43} & 26.84\scalebox{0.7}{$\pm$.30} & 39.15\scalebox{0.7}{$\pm$.35} & 50.24\scalebox{0.7}{$\pm$.45} & 34.05\scalebox{0.7}{$\pm$.31} & 59.88\scalebox{0.7}{$\pm$.36} \\
Dynamic Distill\scalebox{0.7}{++}~\cite{dynamic-distill} & 36.96\scalebox{0.7}{$\pm$.43} & 27.87\scalebox{0.7}{$\pm$.30} & 35.84\scalebox{0.7}{$\pm$.32} & 53.78\scalebox{0.7}{$\pm$.47} & \textbf{37.87\scalebox{0.7}{$\pm$.33}} & 56.23\scalebox{0.7}{$\pm$.35} \\
CDFSL-V~\cite{cdfsl-v} & 38.17\scalebox{0.7}{$\pm$.44} & 26.03\scalebox{0.7}{$\pm$.29} & 39.11\scalebox{0.7}{$\pm$.35} &  53.72\scalebox{0.7}{$\pm$.91} & 35.64\scalebox{0.7}{$\pm$.32} & 58.27\scalebox{0.7}{$\pm$.36} \\
\cmidrule(l){1-7} 
\rowcolor{lightgray}
Ours & \textbf{41.97\scalebox{0.7}{$\pm$.46}} & \textbf{28.34\scalebox{0.7}{$\pm$.30}} & \textbf{51.25\scalebox{0.7}{$\pm$.40}} &  \textbf{58.70\scalebox{0.7}{$\pm$.90}} & 37.80\scalebox{0.7}{$\pm$.46} & \textbf{69.57\scalebox{0.7}{$\pm$.37}} \\
\bottomrule
\end{tabular}}
\label{table: main-acc}
\end{table}

\subsection{Comparison Methods}
\begin{itemize}
    \item \textbf{Random Initialization} is used as our baseline for the experiment. This method entails learning a logistic regression classifier on top of an untrained VideoMAE encoder.
    \item \textbf{VideoMAE}~\cite{videomae} is used as our baseline, whose parameters are initialized with pretrained on the source (Ego4D) and fine-tuned on the support set.
    \item \textbf{STARTUP++}~\cite{startup} is a modified version of STARTUP, used in CDFSL-V for a fair comparison.
    It replaces supervised training during the pretraining stage of the STARTUP with self-supervised pretraining on the source dataset using VideoMAE.
    Subsequently, following STARTUP, it performs the representation learning with the KL-divergence loss and self-supervised contrastive loss~\cite{simclr}.
    \item \textbf{Dynamic Distillation++}~\cite{dynamic-distill} adopts self-supervised training instead of supervised training, similar to the STARTUP++, to ensure a fair comparison.
    Then, it trains the student network with pseudo-labeling while dynamically updating the teacher network's parameters throughout training. 
    This approach involves aligning the student network's predictions for strongly augmented versions of unlabeled target data with the teacher network's weakly augmented counterpart.
    \item \textbf{CDFSL-V}~\cite{cdfsl-v} proposed to use the recent self-supervised technique, VideoMAE~\cite{videomae}, for the pretraining stage, enabling the pretraining with the unlabeled target data to enhance the adaptability to the target domain.
    CDFSL-V also adopts pseudo distillation as a consistency regularization, similar to Dynamic Distillation, during the second training stage.
\end{itemize}
Note that for all comparison methods except random initialization, the weights are initialized from the pretrained model on the Kinetics-400 at the pretraining stage, the same as our initialization.

\begin{table}[tb]
\centering
\caption{\textbf{Inference cost}. Ours adopts the mask ratio $\rho_{\text{infer}} = 0.75$ and ensemble number $P = 2$. Note that existing methods use all tokens from input frames; thus, they have the same values across all metrics. We use a machine equipped with Intel Xeon W-3235 CPU, 128GB RAM, and the NVIDIA Titan RTX GPU to compute the inference cost.}
\centering
\scalebox{0.9}{
\begin{tabular}{lccc}
\toprule
\multirow{2}{*}{Method} & \multirow{2}{*}{{\begin{tabular}{c}Runtime  \\ (ms) \end{tabular}}} & \multirow{2}{*}{GFLOPs} & \multirow{2}{*}{{\begin{tabular}{c}Memory  \\ (MiB) \end{tabular}}}
\\
 & & & \\
\toprule
Random Initialization & \multirow{5}{*}{22.1} & \multirow{5}{*}{68.5} & \multirow{5}{*}{2782} \\
VideoMAE~\cite{videomae} & & & \\
STARTUP++~\cite{startup} & & & \\
Dynamic Distillation++~\cite{dynamic-distill} & & & \\
CDFSL-V~\cite{cdfsl-v} & & & \\
\cmidrule(l){1-4} 
\rowcolor{lightgray}
Ours & \textbf{9.64} & \textbf{37.0} & \textbf{968} \\
\bottomrule
\end{tabular}}
\label{table: main-cost}
\end{table}

\begin{table}[t]
\begin{tabular}{cc}
\begin{minipage}[t]{0.59\textwidth}
\caption{\textbf{Loss component ablation study in the pretraining stage}. \textit{Only reconstruction} is when $\lambda_{{ce}_m}=0$ in \cref{eq:pretrain-loss}. All results are reported with the same mask ratio and ensemble number ($\rho_{\text{infer}} = 0.75$, $P = 2$)}
\centering
\scalebox{0.88}{
\begin{tabular}{lccccc}
\toprule
Method & $\mathcal{L}_{\text{recon}}^{\text{source}}$ & $\mathcal{L}_{\text{recon}}^{\text{target}}$ & $\mathcal{L}_{\text{ce}}^{\text{source}}$ & 1-shot & 5-shot \\
\midrule
Only reconstruction & $\checkmark$ & $\checkmark$ &              & 35.42 & 49.82 \\
Only source         & $\checkmark$ &              & $\checkmark$ & 40.50 & 56.43 \\
Ours                & $\checkmark$ & $\checkmark$ & $\checkmark$ & \textbf{41.97} & \textbf{58.70} \\
\bottomrule
\end{tabular}}
\label{table: loss-pretrain}
\end{minipage}
&
\hspace{0.02\textwidth}
\begin{minipage}[t]{0.39\textwidth}
\centering
\caption{\textbf{Ablation study on the multimodal distillation stage}. \textit{Only RGB Training} row shows the accuracy without distilling the multimodal information at the second training stage. }
\scalebox{0.85}{
\begin{tabular}{lcc}
\toprule
Method & 1-shot & 5-shot \\
\midrule
Only RGB Training & 46.17 & 67.19 \\
RGB+Pose & 49.39 & 67.90 \\
Ours & \textbf{51.25} & \textbf{69.57} \\
\bottomrule
\end{tabular}
}
\label{table: loss-mmdistill}
\end{minipage}
\end{tabular}
\end{table}

\subsection{Quantitative Comparison}
We compare the performance of 5-way 1-shot and 5-shot top-1 action recognition accuracy and inference cost with the prior methods, trained on the Ego4D~\cite{ego4d} dataset as the source domain, on three egocentric datasets as the target domain: EPIC-Kitchens~\cite{epic} in the cooking domain, MECCANO~\cite{meccano} in the industrial-like domain, and WEAR~\cite{wear} in the outdoor workout domain.
We report the accuracy and inference cost of our proposed method with the mask ratio of $\rho_{\text{infer}}$ of $0.75$, using the ensemble number $P = 2$.
\cref{table: main-acc} and \cref{table: main-cost} show that the proposed method consistently outperforms the state-of-the-art CD-FSL with unlabeled target data methods regarding action recognition accuracy and inference cost.

\noindent \textbf{Few-Shot Action Recognition Accuracy}.
Our proposed model outperforms the CDFSL-V by $6.10$ points and the Dynamic Distillation++ by $6.07$ points on the average of three datasets regarding the 5-shot action recognition accuracy.
Significant improvement can be seen in 5-shot accuracy on the WEAR dataset (from $58.28$ to $69.57$) as our method leverages the multiple modalities during training, further mitigating the domain gap between source and target.
Also, our model outperforms the CDFSL-V by $6.12$ points and the Dynamic Distillation++ by $6.95$ points on average of three datasets in the 1-shot setting, even though only one labeled target sample from the support set $S$ is available during the few-shot training stage.

\noindent \textbf{Inference Cost}.
As shown in \cref{table: main-cost}, our approach achieves $2.2$x faster inference speed (ms) than the previous approaches while achieving state-of-the-art few-shot classification performance.
Furthermore, our method reduces 46$\%$ of the theoretical computational cost (GFLOPs), and use only 34$\%$ of memory consumption (MiB) compared to the previous methods.
The existing approaches in \cref{table: main-cost} employ the same architecture (ViT-S) and the same number of input tokens; thus, their inference times remain consistent across implementations.
Our approach enhances efficiency and retains high accuracy, making it particularly advantageous for applications demanding online processing without compromising prediction accuracy.

\begin{wrapfigure}{r}{0.5\linewidth}
\begin{center}
\vspace{-3em}
   \includegraphics[width=\linewidth]{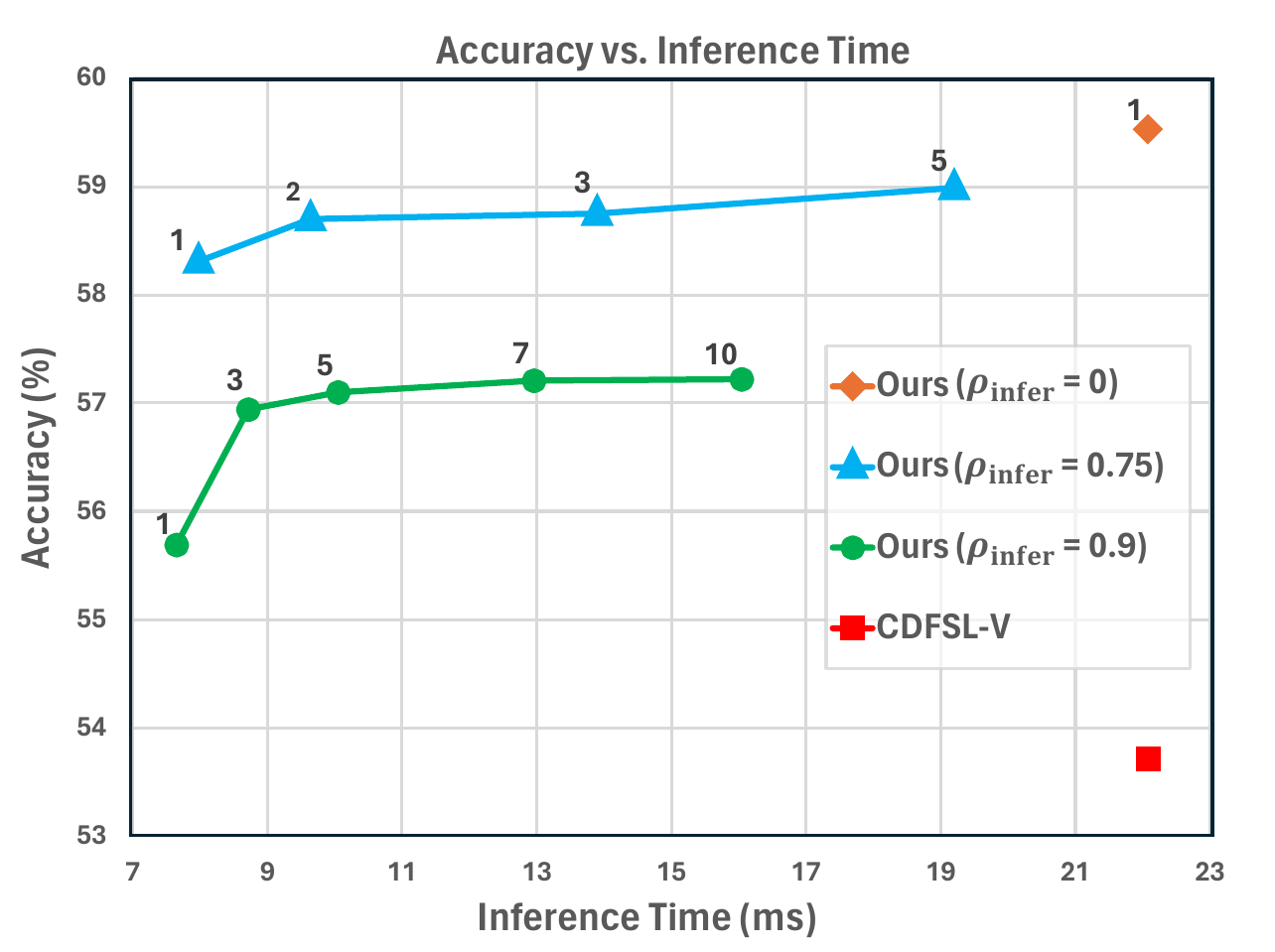}
\end{center}
\vspace{-2em}
   \caption{\textbf{Accuracy vs. inference time}. The trade-off analysis between action recognition accuracy and inference speed is conducted for the existing method and our proposed approach, examining various masking ratios $\rho_{\text{infer}}$ and ensemble numbers $P$. The number near the plots for our proposed method denotes the ensemble number.}
\label{fig: acc-vs-time}
\vspace{-2em}
\end{wrapfigure}

\subsection{Ablation Analysis}
In this section, we conduct the ablation study of the proposed framework. 

\noindent \textbf{Class-discriminative Training in the Pretraining Stage}.
We conduct an ablation analysis on employing the cross-entropy loss during the pretraining stage in \cref{eq:pretrain-loss} to assess its impact on class-discriminative feature learning in conjunction with self-supervised learning.
We report the performance of the best trade-off model, which incorporates multimodal feature distillation to determine the impact of this cross-entropy loss ablation.
As shown in \cref{table: loss-pretrain}, the results verify the importance of cross-entropy loss in obtaining class-discriminative representations, thereby significantly boosting the model's accuracy and reliability for few-shot action recognition in cross-domain scenarios.

\noindent \textbf{Adaptability of the Pre-trained Model}.
Our teacher encoders in the multimodal distillation depend on pre-trained models. The distillation process using target domain data fails if the pre-trained models do not adequately capture the target domain characteristics.
We leverage unlabeled target data during pre-training to adapt the model to the target domain.
As shown in \cref{table: loss-pretrain}, ablation study of the target domain data during the pre-training stage verifies that our method prevents the above issue.

\noindent \textbf{Multimodal Distillation}.
We also conduct an ablation study during the second stage of training to assess the benefits of multimodal feature distillation. 
\cref{table: loss-mmdistill} presents the 1-shot and 5-shot few-shot action recognition accuracy on the WEAR dataset. 
The results confirm that omitting the multimodal distillation stage leads to a decrease in action recognition accuracy. 
This decline underscores the significance of multimodal distillation for enhancing performance in cross-domain few-shot action recognition tasks. 

\noindent \textbf{Accuracy and Speed Trade-off}.
The ablation study examines the impact of the masking ratio $\rho_{\text{infer}}$, applied to input frames during inference, and the ensemble number $P$.
This investigation aims to explore the trade-off between model accuracy and inference speed, providing insights into how variations in $\rho_{\text{infer}}$ and $P$ influence the overall performance and efficiency of the model.

\cref{fig: acc-vs-time} illustrates the accuracy vs. inference time trade-off on the EPIC-Kitchens dataset.
Our most accurate model ($\rho_{\text{infer}}=0, P=1$) significantly surpasses CDFSL-V by $5.23$ points with the same inference speed as the existing method.
Moreover, our fastest model ($\rho_{\text{infer}}=0.9, P=1$) achieves $7.65~\text{ms}$ in inference speed, which is 2.9x faster than the previous method and still outperforms the previous method in terms of accuracy.
Although the fastest model suffers a drop in performance compared to our most accurate model, this performance degradation is modulated by increasing the number of ensemble predictions without sacrificing the inference speed (\eg $P = 3$).

\section{Conclusion}
\noindent\textbf{Conclusion}.
We present MM-CDFSL, the first work to explore the multimodal data for egocentric action recognition in cross-domain and few-shot settings.
We propose training the models for each input modality during the first pretraining stage to gain domain-adapted and class-discriminative features.
Then, we perform multimodal distillation to the student RGB models using teacher models for all modalities to mitigate the domain gap further.
Moreover, we propose ensemble masked inference to reduce the computation cost during inference by masking the input frames while alleviating the drop in performance via ensemble learning.
Experiments on egocentric datasets from three domains, EPIC-Kitchens, MECCANO, and WEAR datasets, demonstrate that our approach outperforms the state-of-the-art CD-FSL with unlabeled target data methods regarding action recognition accuracy and inference speed.

\noindent \textbf{Limitations and future work}.
Our proposed method leverages the optical flow and 2D hand keypoints based on the off-the-shelf optical flow estimator~\cite{flowformer}  and 2D hand keypoints detector~\cite{rtmpose}.
Thus, the bias and errors from the off-the-shelf detector may still affect the input modality information.
In addition, our approach applies constant loss weights for feature distillation losses during the multimodal distillation process, regardless of the specific target dataset. 
This strategy does not account for the varying significance of modalities based on the target domain; for instance, motion information may hold greater relevance than hand pose data in outdoor environments. 
Dynamically adjusting distillation weights according to the modality's relevance in the target domain is crucial for achieving more targeted and efficient training outcomes.
We will leave this for our future efforts.

\section*{Acknowledgements}
This work was supported by JST BOOST, Japan Grant Number JPMJBS2409, and Amano Institute of Technology.

%
%
\bibliographystyle{splncs04}
\bibliography{egbib}

\clearpage

\title{Multimodal Cross-Domain Few-Shot Learning \\ for Egocentric Action Recognition \\ \textmd{-- Supplementary Materials --}}

\titlerunning{MM-CDFSL Suppl.}

\author{Masashi Hatano\inst{1}\orcidlink{0009-0002-7090-6564} \and
Ryo Hachiuma\inst{2}\orcidlink{0000-0001-8274-3710} \and
Ryo Fujii\inst{1}\orcidlink{0000-0002-9115-8414} \and
Hideo Saito\inst{1}\orcidlink{0000-0002-2421-9862}}

\authorrunning{M.~Hatano et al.}

\institute{
Keio University \and
NVIDIA
}

\maketitle

\setcounter{figure}{3}
\setcounter{table}{5}
\begin{center}
    \centering
    \captionsetup{type=figure}
    \includegraphics[width=\linewidth]{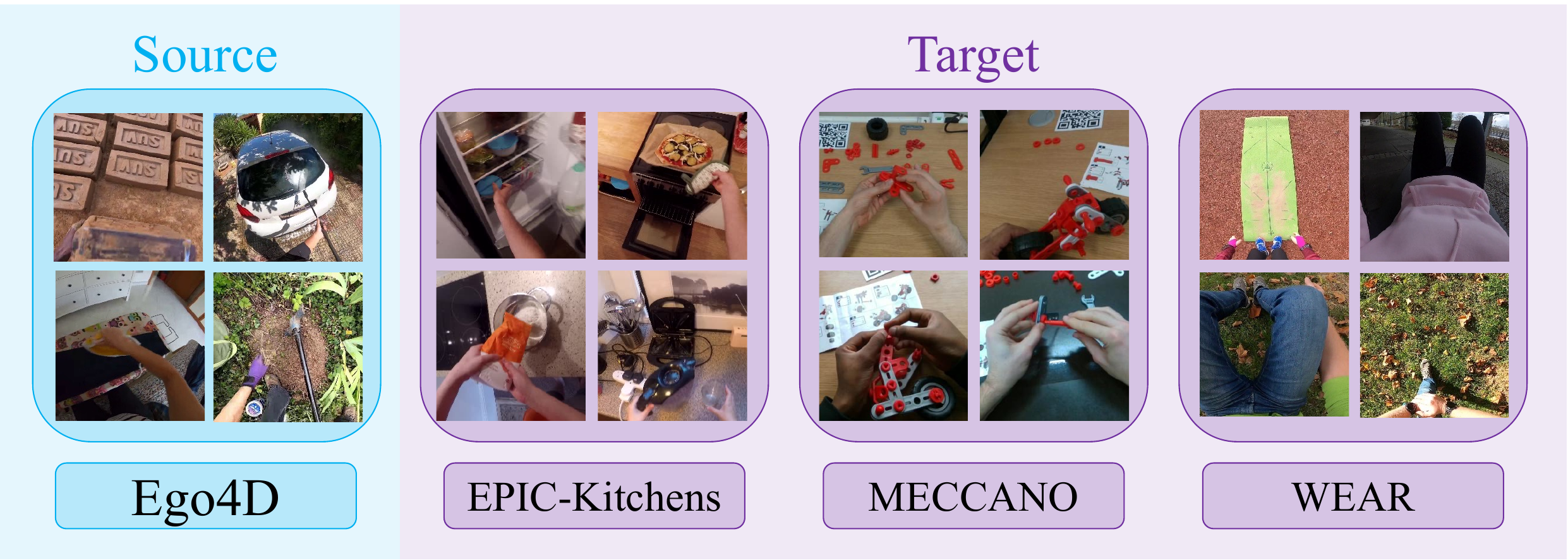}
    \captionof{figure}{\textbf{Samples from each dataset}. A curated selection of RGB images from each dataset showcases the domain gap between the source and target datasets.}
    \label{fig: dataset}
\end{center}

\thispagestyle{empty}
\appendix

\section{Datasets}
In this section, we delve into the datasets used in our study, highlighting the significant domain gap observed in RGB images between the source (Ego4D~\cite{ego4d}) and target (EPIC-Kitchens~\cite{epic}, MECCANO~\cite{meccano}, and WEAR~\cite{wear}) datasets. 
\cref{fig: dataset} presents a curated selection of RGB images drawn from each dataset.
These visual examples underscore the diversity and complexity of cross-domain few-shot learning (CD-FSL) tasks.
\begin{itemize}
    \item \textbf{Ego4D}. The Ego4D dataset comprises a diverse range of activities, including domestic chores in houses, gardening, cleaning, building, cooking, and arts/crafting.
    These actions contain those performed using a single hand, both hands, and various tools, demonstrating the diversity of human activities. Thus, this dataset is well-suited for learning generalizable features required for cross-domain and few-shot settings.
    \item \textbf{EPIC-Kitchens}. The EPIC-Kitchens dataset, serving as the target, is centered around kitchen activities, encompassing actions like ``pouring flour'', ``opening the refrigerator'', ``moving a pizza'', and ``pouring a smoothie'', each composed of a verb-noun pair. Such activities in the kitchens are also present within the Ego4D dataset, our source. However, the EPIC-Kitchens dataset distinguishes itself by its fine-grained action categories, with some actions sharing verbs but differing in nouns and others vice versa. This granular differentiation of actions introduces a challenge for few-shot learning models, necessitating nuanced discernment in action categorization.
    \item \textbf{MECCANO}. The MECCANO dataset comprises detailed recordings of the assembly process for a toy bicycle in an industrial-like scenario. It contains fine-grained actions, including ``aligning a screwdriver to screw'', ``aligning objects'', ``putting a tire'', and ``putting a screw''. The diminutive size of the components poses a substantial challenge for action recognition, requiring exceptional precision to identify and understand the intricate interactions.
    \item \textbf{WEAR}. This dataset captures outdoor workout actions such as ``stretching hamstrings'', ``jogging'', ``pushing-ups'', and ``sitting-ups''. Unlike the other datasets, WEAR focuses on activities not involving hand-object interactions but body movements.
\end{itemize}

\section{Implementation Details}
\noindent \textbf{Experimental Setup}.
We utilize the three modalities: RGB, optical flow, and hand pose. 
For each modality, we select an input sequence comprising $T = 16$ frames sampled at a frequency of $8$ FPS (frames per second). 
Spatial dimensions are standardized at $224 \times 224$ for RGB and optical flow inputs, while hand pose inputs, represented by the heatmap, are resized to $56 \times 56$. 
The number of channels $C_{m}$ are 3, 2, and 21 for RGB, optical flow, and hand pose, respectively. 
We employ FlowFormer~\cite{flowformer} for estimating the optical flow between consecutive frames. 
For the prediction of 2D hand keypoints, which include 21 joints, we use RTMPose~\cite{rtmpose} trained on five public hand pose datasets available through MMPose~\footnote{\url{https://github.com/open-mmlab/mmpose}}.
Subsequently, the Gaussian heatmap of size $56 \times 56$ is produced, with a standard deviation $\sigma$ set to 10.
Following the VideoMAE~\cite{videomae} experimental setup, we adopt the mask ratio $\rho_{\text{pretrain}}$ of $0.9$ during our pretraining stage.
In addition, we use the mask ratio $\rho_{\text{distill}} = 0.75$ during the multimodal distillation for all experiments.
We use a machine equipped with Intel Xeon W-3235 CPU, 128GB RAM, and the NVIDIA Titan RTX GPU to compute the inference speed.

\noindent \textbf{Training}.
In the pretraining stage, we train the model, which consists of ViT-S models, pretrained on the Kinetics-400 dataset~\cite{kinetics-400} and a classifier head, which is attached to the ViT-S backbone, for all modalities for 100 epochs. For training settings, we generally follow the VideoMAE~\cite{videomae}.
During the multimodal distillation stage, we train the student RGB model for 100 epochs using the AdamW optimizer~\cite{adamw}, with a peak learning rate of $2e-3$, linearly increased for the first $10$ epochs of the training and decreased to $1e-6$ until the end of training with cosine decay~\cite{cosine-decay}.
Note that we linearly scaled the peak learning rate with respect to the overall batch size. Regarding the parameters for the loss function, we empirically adapt the balancing hyperparameters $\lambda_{\text{ce}_{m}}$ to $5e-2$ for RGB, and $1e-2$ for optical flow and hand pose input modality. 

\noindent \textbf{Evaluation Metrics}.
Following the existing CD-FSL work, we report the top-1 accuracy on the query set $Q$ in the target validation set over 600 runs to measure action recognition performance.
To benchmark efficiency, we quantify the model's performance by measuring the forward pass time during inference.
We report the inference time averaged over $600$ iterations.

\section{Visualization of Feature Representations}
\begin{figure}[t]
\begin{center}
   \includegraphics[width=\linewidth]{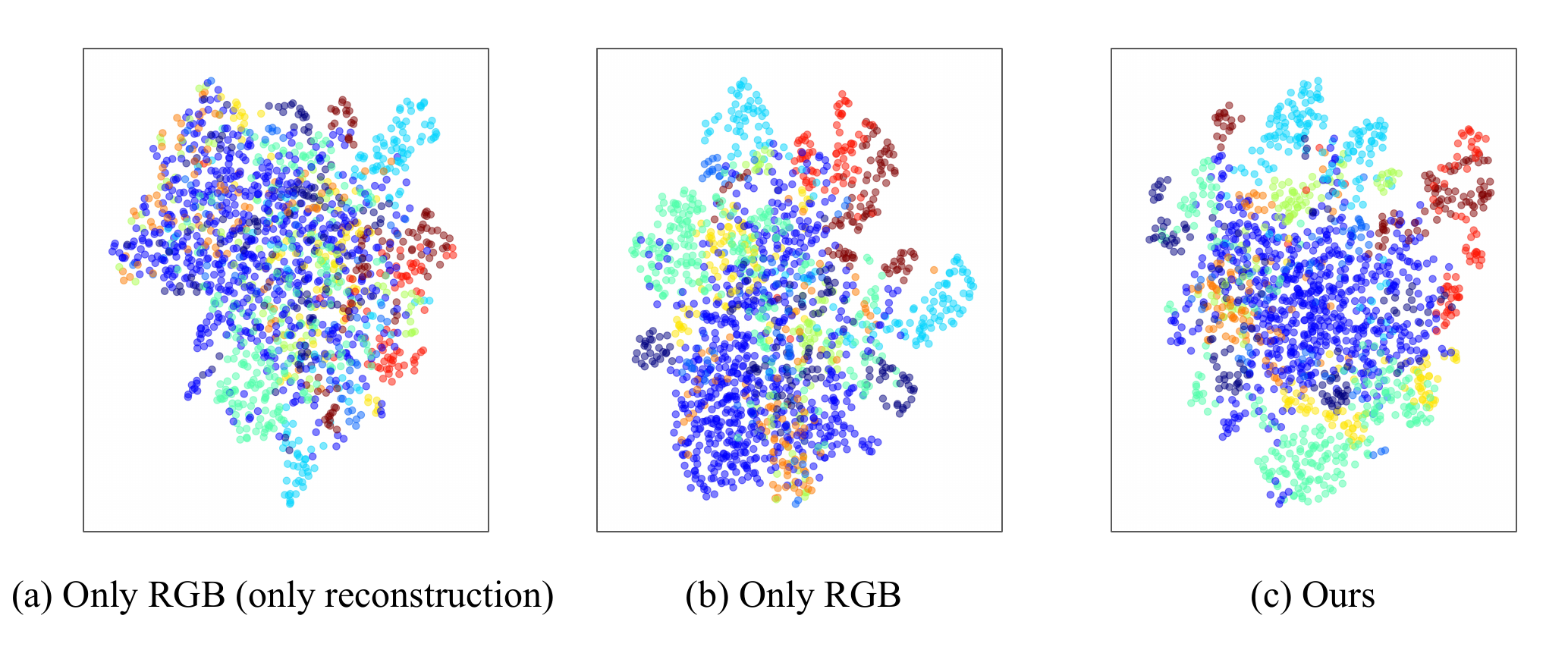}
\end{center}
   \caption{\textbf{Comparative UMAP visualization of feature representations}. UMAP plot of 10 classes from EPIC-Kitchens validation set with features obtained from (a) Only RGB (only reconstruction), (b) Only RGB, and (c) Ours.}
\label{fig: umap}
\end{figure}

We compare the class-discriminativeness of embeddings extracted from three encoders: Only RGB (only reconstruction), Only RGB, and Ours.
The only RGB (only reconstruction) model is trained without multimodal distillation, and $\lambda_{\text{ce}_{\scalebox{0.5}{RGB}}} = 0$ is used during the pretraining stage.
The only RGB model is trained without multimodal distillation, and $\lambda_{\text{ce}_{\scalebox{0.5}{RGB}}} = 0.05$ is used during the pretraining stage.
\cref{fig: umap} shows the UMAP plot of 10 classes from EPIC-Kitchens datasets.
We see that only RGB model creates better grouping on the embeddings of the target datasets than only RGB (only reconstruction) model.
This result supports that using the cross-entropy loss helps learn the class-discriminative features during the pretraining stage.
We further see that the multimodal distillation also helps learn discriminative features compared to the only RGB model.
 
\section{Loss Weight Ablation}
\begin{table}[tb]
\centering
\caption{\textbf{Loss wight ablation}. We conduct an ablation study on the loss weight for cross-entropy loss on RGB modality pertaining.}
\centering
\scalebox{1}{
\begin{tabular}{lcccc}
\toprule
$\lambda_{\text{ce}_{\scalebox{0.5}{RGB}}}$ & 1 & 0.1 & 0.05 & 0.01\\
\toprule
5-shot & 54.58 & 56.68 & \textbf{57.07} & 52.40 \\
\bottomrule
\end{tabular}}
\label{table: loss-weight}
\end{table}

We present an ablation study focused on the impact of adjusting the loss weight for the cross-entropy loss on the RGB modality $\lambda_{\text{ce}_{\text{RGB}}}$ during the pretraining stage. 
The loss weight for the cross-entropy loss $\lambda_{\text{ce}_{\text{RGB}}}$ serves as a critical hyperparameter that balances the contribution of the cross-entropy loss to the total loss function. 
For the ablation study, we varied the value of the loss weight across a predefined range $\lambda_{\text{ce}_{\text{RGB}}} \in \{1, 0.1, 0.05, 0.01 \}$.
We report the 5-way 5-shot action recognition accuracy on the EPIC-Kitchens dataset of the only RGB model with varied hyperparameter $\lambda_{\text{ce}_{\text{RGB}}}$ in \cref{table: loss-weight}.
Our ablation analysis reveals a critical insight: a high or low loss weight for the RGB modality, $\lambda_{\text{ce}{\text{RGB}}}$, during the pretraining stage can detrimentally affect the acquisition of class-discriminativeness on the target data.
Assigning a high loss weight to the cross-entropy reduces the relative contribution of learning from unlabeled target data. 
Conversely, a low loss weight fails to capture class discriminativeness adequately.


\end{document}